\documentclass{article}
\usepackage{ijcai20}

\usepackage{times}
\usepackage{soul}
\usepackage{url}
\usepackage[hidelinks]{hyperref}
\usepackage[utf8]{inputenc}
\usepackage[small]{caption}
\usepackage{graphicx}
\usepackage{amsmath}
\usepackage{amsthm}
\usepackage{booktabs}
\usepackage{algorithm}
\usepackage{algorithmic}
\usepackage{multirow}
\usepackage{multicol}
\usepackage{lipsum}
\usepackage{tabularx}
\newcommand{\tabincell}[2]{\begin{tabular}{@{}#1@{}}#2\end{tabular}}
\usepackage{amsmath}
\usepackage{color}
\usepackage{bm}
\usepackage{subfigure} 
\usepackage{graphicx}

\newcommand{\citet}[1]{\citeauthor{#1} \shortcite{#1}}
\newcommand{\citep}{\cite}

\title{Domain Adaptation for Semantic Parsing}

\author{
Zechang Li$^{1,2}$\and
Yuxuan Lai$^1$\and
Yansong Feng$^{1,3}$\footnote{Corresponding Author}\And
Dongyan Zhao$^{1,2}$\\
\affiliations
$^1$Wangxuan Institute of Computer Technology, Peking University, Beijing, China\\
$^2$Center for Data Science, Peking University, Beijing, China\\
$^3$The MOE Key Laboratory of Computational Linguistics, Peking University, China\\
\emails
\{zcli18, erutan, fengyansong, zhaody\}@pku.edu.cn
}

\begin{document}

\maketitle

\newcommand\blfootnote[1]{%
	\begingroup
	\renewcommand\thefootnote{}\footnote{#1}%
	\addtocounter{footnote}{-1}%
	\endgroup
}

\begin{abstract}

Recently, semantic parsing has attracted much attention in the community. Although many neural modeling efforts have greatly improved the performance, it still suffers from the data scarcity issue. In this paper, 
we propose a novel semantic parser for domain adaptation,
where we have much fewer annotated data in the target domain compared to the source domain. Our semantic parser benefits from a two-stage coarse-to-fine framework, thus can provide different and accurate treatments for the two stages, i.e., focusing on domain invariant and domain specific information, respectively. In the coarse stage, our novel domain discrimination component and domain relevance attention encourage the model to learn transferable domain general structures. In the fine stage, the model is guided to concentrate on domain related details. Experiments on a benchmark dataset show that our method consistently outperforms several popular domain adaptation strategies.  Additionally, we show that our model can well exploit limited target data to capture the difference between the source and target domain, even when the target domain has far fewer training instances.

\end{abstract}
\section{Introduction}

\blfootnote{Copyright International Joint Conferences on Artificial Intelligence (\href{http://ijcai.org/}{IJCAI.ORG}). All rights reserved.}

Semantic parsing is the task of transforming natural language utterances into meaning representations such as executable structured queries or logical forms. Despite traditional syntactic parsing style models, there have been many recent efforts devoted to end-to-end neural models in a supervised manner~\cite{dong2016language,sun2018semantic,bogin2019global}. It is known that such models usually require many labeled data for training and are often hard to transfer to new domains, since the meaning representations may vary greatly between different domains, e.g., the \texttt{calendar} and \texttt{housing} domains share less similarity in their meaning representations~\cite{wang2015building}. 

However, there has been relatively less attention to the domain adaptation for semantic parsing. This is not an easy task, since one has to deal with the transfer of semantic representations, including both structural levels and lexical levels. And it is often more challenging than the transfer of a sentence classification model. Moreover, contrast to other conventional domain transfer tasks, e.g., sentiment analysis, where all labels have been seen in source domains, semantic parsing models are expected to generate domain specific labels or tokens with limited target domain annotations, e.g., {\small\textsf{attendee}} only appears in the \texttt{calendar} domain. 
These observations suggest that more efforts are required to deal with the query structure transfer and few-shot token generation issues when we perform domain adaptation for semantic parsing. 

\begin{table}
\centering
\footnotesize
\begin{tabular}{p{1.2cm}c}
\toprule
\textbf{Domain} & \textbf{Instance} \\ \midrule
\texttt{calendar} & \tabincell{p{6.5cm}}{
{utterance}:~\textit{meetings attended by two or more people} \\ 
{logical form}:~{\scriptsize\textsf{\textbf{listValue (countComparative (getProperty (singleton} en.meeting \textbf{) ( string !type ) ) ( string} attendee \textbf{) ( string \textgreater= ) ( number }2\textbf{ ) ) }}}} \\ \midrule
\texttt{housing} & \tabincell{p{6.5cm}}{
{utterance}:~\textit{housing units with 2 neighborhoods} \\
{logical form}:~{\scriptsize\textsf{\textbf{listValue (countComparative (getProperty (singleton} en.housing\_unit \textbf{) ( string !type ) ) ( string} neighborhood \textbf{) (string = ) (number }2\textbf{ ) ) }}}} \\
\bottomrule
\end{tabular}
\caption{Examples of paired utterances and their logical forms from the OVERNIGHT dataset. The {\scriptsize\textsf{\textbf{bold}}} tokens in logical forms are usually domain invariant, which can be seen as patterns generalized across different domains.}
\label{tab:sketch}
\end{table}

An intuitive solution to solve this problem is to build a two-stage model, where a coarse level component focuses on learning more general, domain invariant representations, and a fine level component should concentrate on more detailed, domain specific representations. Take the two utterances in Table~\ref{tab:sketch} as an example. Although they come from different domains, they both express the comparison between certain properties and values, querying certain types of entities ({\small\textsf{meeting}} or {\small\textsf{housing unit}}), with several properties ({\small\textsf{attendee}} or {\small\textsf{neighborhood}}) specified  ({\small\textsf{\textgreater = 2}} or {\small\textsf{= 2}}).
We can see that the \textsc{comparative} pattern tends to be domain invariant and can be more easily transferred in the coarse level, while domain related tokens, e.g., the category and property names, should be concentrated in the fine stage. 

In this work, we propose a novel two-stage semantic parsing approach for domain adaptation.
Our approach is inspired by the recent coarse-to-fine (coarse2fine) architecture~\cite{dong2018coarse}, where the coarse step produces general intermediate representations, i.e., sketches, and then the fine step generates detailed tokens or labels. 

However, the coarse2fine architecture can not be applied to domain adaptation directly, because there is no guarantee for the two stages to achieve our expected different purposes, since the predicate-only intermediate sketch can just provide a distant signal. We thus propose two novel mechanisms, an adversarial domain discrimination  and a domain relevance attention to enhance the encoders and decoders, respectively. They drive the model to learn domain general and domain related representations in different stages, and help to focus on different clues during decoding. We conduct experiments on the OVERNIGHT dataset~\cite{wang2015building}, and outperform conventional semantic parsing and popular domain transfer methods. Further analysis shows that 
both adversarial domain discrimination and domain relevance attention can make the most of the coarse-to-fine architecture for domain adaptation.

Our contributions are summarized as follows:

$\bullet$ We propose a novel two-stage semantic parsing model for domain adaptation, where the coarse step transfers the domain general structural patterns and the fine step focuses on the difference between domains.

$\bullet$ We design two novel mechanisms, adversarial domain discrimination  and domain relevance attention to enhance the encoders and decoders, which help the model to learn domain invariant patterns in the coarse stage, while  focusing on domain related details in the fine stage.

% $\bullet$ Empirical results on OVERNIGHT demonstrate that our approach outperforms several popular semantic parsing and domain adaptation methods. 
% Our approach is also robust with various amounts of target domain training instances.

\section{Task Definition}

Formally, given a natural language utterance $X = x_1, ..., x_{|X|}$ with length $|X|$, the semantic parsing task aims at generating a logical form $Y = y_1, ..., y_{|Y|}$ with length $|Y|$, which formally presents the meaning of $X$, but in predefined grammar. In the domain adaptation settings, each instance $(x_i, y_i)$ is also associated with a specific domain, e.g., \texttt{housing} or \texttt{calendar}, etc. Specifically, domains with sufficient labeled instances are treated as source domains $\mathcal{D}_{S_1}, ..., \mathcal{D}_{S_k}$. And if a domain include far less labeled instances than any source domains, we treat it as  a target domain $\mathcal{D}_T$, i.e., $|\mathcal{D}_{S_i}| >> |\mathcal{D}_{T}|, \forall i$. We denote the combination of source domains as $\mathcal{D}_{S}$. Our goal is to learn a semantic parser for the target domain by exploring both abundant source domain data and limited target domain annotations.

\section{DAMP}
\label{c2fmodel}

We propose a Domain-Aware seMantic Parser, DAMP, within the coarse2fine framework~\cite{dong2018coarse}, which introduces an
intermediate sketch ($A = a_1, ...,a_{|A|}$) to bridge natural language utterances and logical forms.
The procedures to generate sketches and logical forms are called the coarse stage and fine stage, respectively.
Our main idea is to disentangle the domain invariant sketches and domain specific tokens in the two stages, respectively.

\begin{figure*}
\centering
\includegraphics[width=0.8\textwidth]{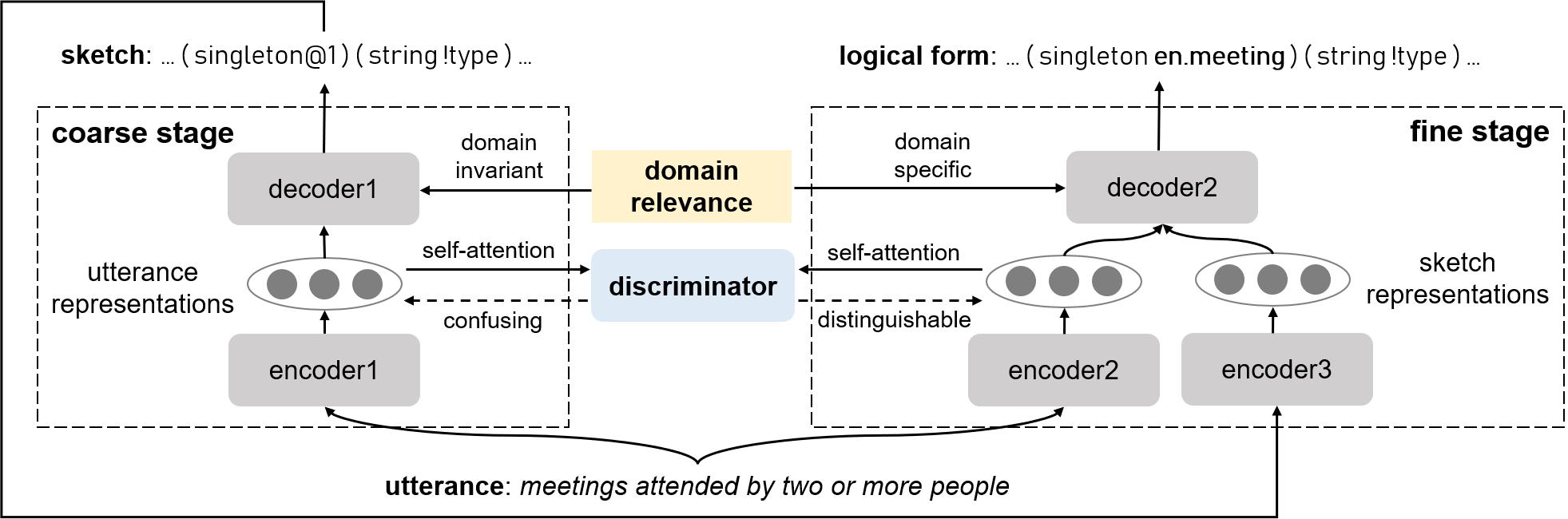}
\caption{Overview of DAMP. The left part is the coarse stage and the right shows the fine stage. The blue module in the middle is the domain discrimination component while the yellow shows the domain relevance attention. }
\label{fig:model}
\end{figure*}

However, it is not appropriate to directly apply the vanilla coarse2fine model to the domain adaptation scenario, since it does not explicitly consider domain information in designing either sketch or model architectures. To alleviate this problem, we first approximate logical form tokens shared by more than 50\% source domains as sketch tokens, since we assume sketches are domain general and should be shared across different domains. The rest tokens are regarded as domain related and should be generated in the fine stage.
We also introduce multi-task based domain discrimination and domain relevance attention to the encoder and decoder procedures, encouraging the parser to focus on different aspects, i.e. domain general and domain specific, during the coarse and fine stages, respectively. The overview of DAMP is illustrated in Figure~\ref{fig:model}. The implemention is open source.\footnote{https://github.com/zechagl/DAMP}

In the coarse stage, utterance representations $\mathbf{U}^c=\{\mathbf{u}_k^c\}_{k=1}^{|X|}$ are produced by $encoder1$ given the utterance $X$. Afterwards, $\mathbf{U}^c$ are fed into $decoder1$ via attention mechanism to generate the sketch A.

In the fine stage, to capture the utterance information in different aspects, we adopt another encoder, $encoder2$, and the new utterance representations are $\mathbf{U}^f=\{\mathbf{u}_k^f\}_{k=1}^{|X|}$.
There is also $encoder3$ to encode the sketch into sketch representations $\mathbf{S}^f = \{\mathbf{s}_k^f\}_{k=1}^{|A|}$.
$decoder2$ takes $\mathbf{U}^f$ and $\mathbf{S}^f$ with attention mechanism and generate the final logical form $Y$.

\subsection{Encoder: Domain Discrimination}
\label{Domain_Discrimination}

In order to constrain the utterance encoders in the coarse and fine stages to focus on domain invariant and domain specific information, respectively, we adopt a domain discrimination component over $\mathbf{U}^c$ and $\mathbf{U}^f$. This can guide $\mathbf{U}^c$ more consistent among domains, while maintaining $\mathbf{U}^f$ distinguishable in the fine stage.

Specifically, in the \textbf{coarse stage}, the utterance representations are aggregated via self-attention as: 
\begin{equation}
\mathbf{u}^{c} = \mathbf{U}^{c} \cdot {\bm \alpha}_e^c, ~~~~ {\bm \alpha}_e^c = softmax(\mathbf{U}^c \cdot \mathbf{w}^c_{\alpha e})
\end{equation}
where $\mathbf{w}^c_{\alpha e}$ is a trainable parameter. The domain discriminator further computes $ p^c = \sigma(\mathbf{w}^c_d \mathbf{u}^c + \mathbf{b}^c_d) $, which is the probability of the utterance comes from the source domains. 
The $\mathbf{w}^c_d$ and $\mathbf{b}^c_d$ are parameters and $\sigma$ stands for sigmoid function.
To guide the model confusing among the domains, we perform gradient ascent over the negative log-likelihood of $p^c$. The corresponding loss function for a gradient descent optimizer is (notice that the minus sign is removed for gradient ascent):
\begin{equation}
    L_c^D = \frac{1}{|\mathcal{D}|} (\sum_{X, Y \in \mathcal{D}_S} \log p^c + \sum_{X, Y \in \mathcal{D}_T}  \log (1-p^c))
\end{equation}

In the \textbf{fine stage}, we obtain  
$p^f$, the probability of the utterance coming from source domains, based on $\mathbf{U}^f$.
But here, our target is to make it more discriminative. Thus a conventional gradient descent is adopted, and the corresponding loss function is:
\begin{equation}
    L_f^D = - \frac{1}{|\mathcal{D}|} (\sum_{X, Y \in \mathcal{D}_S} \log p^f + \sum_{X, Y \in \mathcal{D}_T}  \log (1-p^f))
\end{equation}

\subsection{Decoder: Domain Relevance Attention}

We observe that there are many words useful to determine the patterns of sketches, while others are more likely to associate with the domain specific tokens in the logical forms.
Consider the first example in Table~\ref{tab:sketch}, the domain general words like \textit{by two or more} are associated with the comparison sketch in the coarse stage. 
On the other hand, domain related tokens like \textit{meetings} and \textit{attended} help us fill the missing entities and properties during the fine stage. 
Therefore, we propose a domain relevance attention mechanism to integrate this prior to the decoding procedure.

Formally, in the time step $t$ of the \textbf{coarse stage}, the predicated distribution is:
\begin{align}
P(a|a_{<t}, X) &= softmax(FNN([\mathbf{d}_t; \mathbf{c}_t; \mathbf{c}_t^{pri}]))  \\
\mathbf{d}_t &= LSTM(\mathbf{i}_{t-1}, \mathbf{d}_{t-1})
\end{align}
where $[\cdot; \cdot]$ denotes vector concatenation, $a_{t-1}$ is the $t-1$th token in the sketch, $\mathbf{i}_{t-1}$ is the word embedding of $a_{t-1}$, $\mathbf{d}_t$ is the hidden state of $t$th step of the decoder LSTM, and $FNN$ is a two-layer feed-forward neural network. $\mathbf{c_t}$ and $\mathbf{c_t^{pri}}$ are the context representations 
% which is the attention results, 
computed as:
\begin{align}
    \mathbf{c}_t &= \mathbf{U}^c \cdot {\bm \alpha}_{t}, 
    {\bm \alpha}_{t} = softmax(\mathbf{U}^c \cdot \mathbf{d}_t) \\
    \mathbf{c}_t^{pri} &= \mathbf{U}^c \cdot {\bm \alpha}_{t}^{pri},
    {\bm \alpha}_{t}^{pri} = softmax((\mathbf{U}^c \cdot \mathbf{d}_t) \circ \mathbf{q}^c)
\end{align}
where $\circ$ stands for the element-wise multiply, and $\mathbf{q^c}$ is a vector of length $|X|$.
% , indicating this utterance's domain, $\mathcal{D}$. 
The $k$th dimension of $\mathbf{q^c}$, i.e.,~$\mathbf{q^c_k}$, is 1 if $x_k$ is relevant to this utterance's domain $\mathcal{D}$, or $r^c$ otherwise. Since the hyper-parameter $r^c >> 1$, this will guide the decoder to attend to the domain invariant words in the utterance, which facilitates the generation of domain invariant sketch.

To determine whether a word $x_k$ is relevant to the domain $\mathcal{D}$, we use the cosine similarity between the word and the name of the domain via word embeddings. We assume that each utterance usually mentions domain specific information within a relatively general pattern. Thus, a few words with top similarities are considered as domain relevant in each utterance, and others are recognized as domain invariant.

%In the \textbf{fine stage}, the procedure is similar, but has two different points. Firstly, taking both the sketch and the utterance as input, the input vector $\mathbf{i}_t$ shares the  sketch encoder $\mathbf{s}^f_k$, if the last decoder output $y_{t-1}$ corresponds to the sketch token $a_k$, or the word embedding of $y_{t-1}$ otherwise:

In the \textbf{fine stage}, the domain relevance attention works similarly to that in the coarse stage, but with two differences.

Firstly, the input to the decoder LSTM, $\mathbf{i}_t$, is not always the embedding of the previous decoder output $y_{t-1}$. If $y_{t-1}$ corresponds to the sketch token $a_k$, $\mathbf{i}_t$ will be switched to  $\mathbf{s}^f_k$:
\begin{align}
    \mathbf{i}_t &= \begin{cases}
    \mathbf{s}_k^f & y_{t-1}\text{~corresponds~to~}a_k \\
    embedding(y_{t-1}) & \text{otherwise}
    \end{cases}
\end{align}

Second, in the fine stage, the model is expected to focus on domain specific tokens. Therefore, for the prior vector, $\mathbf{q}^f_k$ is set to 1 if $x_k$ is irrelevant to domain $\mathcal{D}$, and it will be set to $r^f$ for the rest cases, which is contrary to the coarse stage.

\subsection{Training and Inference}

The final loss functions for both the coarse stage (i.e., sketch generation) and the fine stage (i.e., logical form generation) are the linear combinations of original cross-entropy loss and the domain discrimination loss (See Sec.~\ref{Domain_Discrimination}):
\begin{align}
    L_c &= \lambda_c L_c^D -\frac{1}{|\mathcal{D}|}\sum_{X, Y \in \mathcal{D}} \sum_{t} \log p(a|a_{<t}, X) \\
    L_f &= \lambda_f L_f^D -\frac{1}{|\mathcal{D}|}\sum_{X, Y \in \mathcal{D}} \sum_{t} \log p(y|y_{<t}, A, X)
\end{align}
where $\lambda_c$ and $\lambda_f$ are hyper-parameters to  trade off between the two loss terms.

For the inference procedure, we first acquire the sketch via $\hat{A} = \arg\max_{A'}p(A'|X)$. Then $\hat{A}$ is adopted to predict the logical form as $\hat{Y} = \arg\max_{Y'}p(Y'|\hat{A}, X)$. We use beam search for decoding where the beam size is 3.

\section{Experiments}
Our experiments are designed to answer the following questions:  (1) Whether our model can deal with domain adaptation? (2) Whether our domain discrimination can help encoders learn domain invariant and specific representations, respectively? (3) Whether our domain relevance attention can help decoders generate rare domain-specific tokens? (4) How our model performs in more tough target domain settings?

\subsection{Setup}

We conduct experiments on OVERNIGHT~\cite{wang2015building}, a semantic parsing dataset with paired utterances and logical forms (in function calling styled Lambda-DCS) in 8 domains, 4 of which are relatively small, i.e., \texttt{publications}, \texttt{calendar}, \texttt{housing} and \texttt{recipes}. For domain adaptation settings, each time, we treat one of the 4 small domains as the target domain, and consider the rest 7 domains as the source domain. Only 10\% of the target domain training data is employed, to stimulate a harsher domain adaptation scenario. We randomly sample 20\% of training instances for validation. Detailed statistics are shown in Table~\ref{tab:overnight}. We evaluate the model performance with the widely adopted exact match rate (\textbf{EM})~\cite{dong2018coarse,kennardi2019domain}.

\begin{table}
\centering
\footnotesize
\begin{tabular}{lrrc}
\toprule
\textbf{Domain} & \textbf{Train} & \textbf{Dev} & \textbf{Test} \\
\midrule
\texttt{publications}  & 512   & 128 & 161  \\
\texttt{calendar}      & 535   & 134 & 168  \\
\texttt{housing}       & 601   & 151 & 189  \\
\texttt{recipes}       & 691   & 173 & 216  \\
\texttt{restaurants}   & 1060  & 265 & -  \\
\texttt{basketball}    & 1248  & 313 & -  \\
\texttt{blocks}        & 1276  & 320 & -  \\
\texttt{socialnetwork} & 2828  & 707 & -  \\
\bottomrule
\end{tabular}
\caption{Detailed statistics of the OVERNIGHT dataset.}
\label{tab:overnight}
\end{table}

For implementation details, word embeddings are initialized with Glove~\cite{pennington2014glove}. And our \textit{encoder}s are Bi-LSTM with 300 hidden size. For domain relevance attention, $r^c = 60$ and  $r^f = 2$. As for balance parameters, $\lambda_c$ is 0.4 and $\lambda_f$ is 0.2. All activation functions are \textit{tanh}. The dropout rate and L2 regularization rate are $0.6$ and $1e-5$, respectively, with batch size of 64. We use the RMSProp optimizer \cite{tieleman2012lecture} with learning rate$=1e-3$ and decay factor $=0.9$.

We compare our DAMP with two widely used semantic parsing baselines following a simple data mixing adaptation strategy, where we mix training data from both source and target domains, and validate on the target domain.

\textbf{\citet{sutskever2014sequence}} is the basic seq2seq model with attention, which generates logical forms from utterances directly.

\textbf{\citet{dong2018coarse}} is the  conventional coarse2fine semantic parsing model.

We also apply three popular domain transfer methods upon the coarse2fine architecture,\footnote{All three transfer methods are adapted to our semantic parsing task according to our implementation.} including:

\textbf{\citet{liu2016recurrent}} is a parameter sharing approach, where 
{we share the parameters in utterance to sketch procedure}, i.e. the coarse stage, and train the fine state merely with target domain data.

\textbf{\citet{kennardi2019domain}} is a pretraining adaptation method. We adapt it in our case by first training coarse2fine with all source domain data, and then fine-tuning with target domain data only.

\textbf{\citet{ganin2015unsupervised}} is a widely used adversarial training method. We adapt it to our task by introducing adversarial losses to the encoders in both coarse and fine stages.

\subsection{Main Results}

\begin{table*}
\centering
\footnotesize
\begin{tabular}{lrrrrr}
\toprule
  & \texttt{recipes} & \texttt{publications} & \texttt{calendar} & \texttt{housing} & {\bf average} \\
\midrule
\citet{sutskever2014sequence} & 58.80 & 36.64 & 34.52 & 36.50 & 41.62 \\
\citet{dong2018coarse} & 62.96 & 38.51 & 38.10 & 39.15 & 44.68 \\
\midrule
\citet{liu2016recurrent} & 51.39 & 27.33 & 27.97 & 33.86 & 35.14 \\
\citet{kennardi2019domain} & 59.72 & 40.99 & 43.45 & 42.32 & 46.62 \\
\citet{ganin2015unsupervised} & 68.06 & 40.37 & \textbf{44.04} & 41.27 & 48.44 \\
\midrule
DAMP & \textbf{72.22} & \textbf{45.96} & 39.88 & \textbf{43.39} & \textbf{50.36} \\
\bottomrule
\end{tabular}
\caption{Performance  of different models on OVERNIGHT. The evaluation metric is EM of the final logical forms.}
\label{tab:da}
\end{table*}

We summarize all the model performance in  Table~\ref{tab:da}. We can see that our DAMP  outperforms other methods in almost all domains, delivering the best overall performance.

It is not surprising that  \citet{dong2018coarse} performs better than \citet{sutskever2014sequence} in all domains. But, with the simple data mixing strategy,  both of them perform worse compared to other models. The reasons could be that simply mixing data from the source and target domains together may confuse the models. Although coarse2fine has two layers to take different treatments for coarse patterns and fine details, there are no explicit mechanisms to help the model to learn general structural patterns from source domains or to focus on domain specific tokens from the target domain.

When applying popular domain adaptation methods to coarse2fine, we can see that 
\citet{liu2016recurrent} is even 9\% worse than vanilla coarse2fine. The main reason may be that only a small number of target domain data are utilized in the fine stage, which is not sufficient to  teach the model  to properly fill domain related tokens into domain generic sketches. Comparing with the coarse2fine model, its EM score drops from 72.7\% to 56.8\% on average, given the golden sketches. \citet{kennardi2019domain} performs better than the vanilla coarse2fine. Compared to \citet{liu2016recurrent}, it conducts target domain fine-tuning in the fine stage. But it still performs worse than our DAMP, since DAMP can learn domain general patterns and domain specific details at different stages in an explicit way, with about 4\% improvement in EM. In our implementation, \citet{ganin2015unsupervised} use adversarial losses to produce better  representations at both coarse and fine stages, and performs slightly better than \citet{kennardi2019domain}. Our DAMP also benefits from the domain relevance attention, thus can perform appropriate decoding in different stages. We also notice that our DAMP performs the best in three domains, except \texttt{calendar}. After examining the cases, we find the reasons may be that our way to compute domain relevance is oversimplified, and most of the domain related keywords are not that similar to \texttt{calendar}. We believe a more sophisticated relevance model can definitely improve the performance, which we leave for future work.

\paragraph{Ablation study.} We perform an ablation study by taking \textit{recipes} as the target domain. Despite EM of the final logical form, we also introduce EM of the obtained sketch and EM of the logical form with oracle sketch as evaluation metrics. As shown in Table~\ref{tab:ablation}, both $\text{DAMP}{-dis}$ and $\text{DAMP}{-att}$ outperform the vanilla coarse2fine model. And the main improvement comes in the coarse stage, showing that our two  mechanisms can  well leverage domain invariant features across multiple domains.
Compared to $\text{DAMP}{-att}$, $\text{DAMP}{-dis}$ performs better, specifically with about 2.8\% improvement in the fine stage (LF$_{\text{oracle}}$). This indicates our domain relevance component is more beneficial for domain adaptation, especially for the target domain transfer in the fine stage.

\begin{table}
\centering
\footnotesize
\begin{tabular}{lrrc}
\toprule
Model                       & Sketch & LF$_{\text{oracle}}$ & LF \\
\midrule
DAMP & 83.80 & 85.19 & 72.22 \\
$\text{DAMP}{-dis}$ & 82.87 & 85.19 & 70.83 \\
$\text{DAMP}{-att}$ & 81.94 & 82.41 & 68.06 \\ \midrule
coarse2fine & 73.61 & 82.87 & 62.96 \\
\bottomrule
\end{tabular}
\caption{The results of the ablation study, where we report the EM rate of intermediate sketch (Sketch), logical form with oracle sketch (LF$_{\text{oracle}}$), and the final logical form (LF). $\text{DAMP}{-dis}$ is the main model without the domain discrimination component and $\text{DAMP}{-att}$ is without the domain relevance attention (conventional attention only). }
\label{tab:ablation}
\end{table}

\subsection{Domain Discrimination}

\begin{figure}
\centering
\hbox{\includegraphics[width=0.47\textwidth]{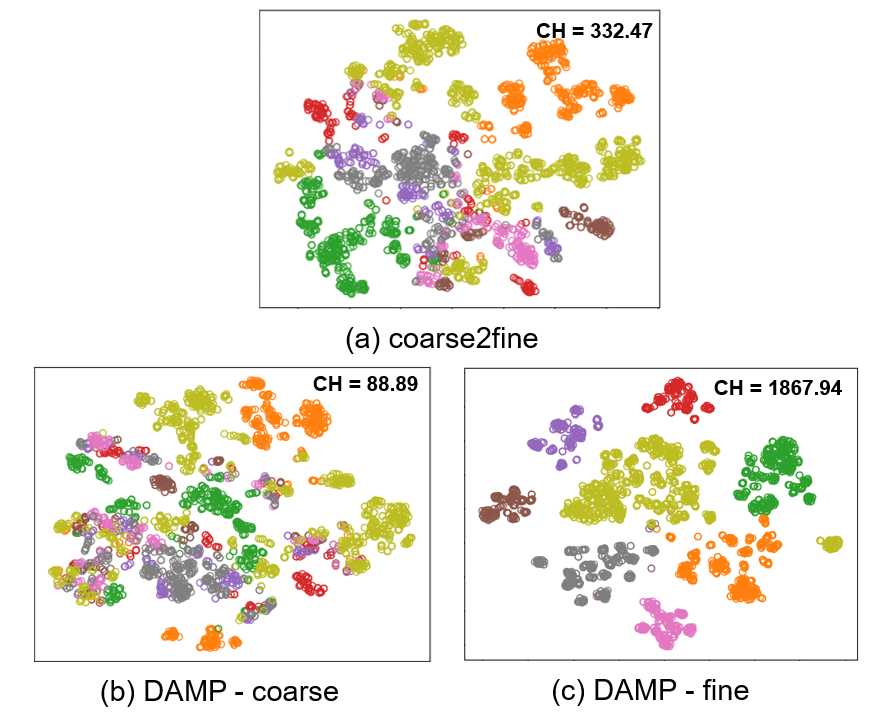}}
\caption{t-SNE visualization of utterance representations (i.e. $\mathbf{U}^c$ and $\mathbf{U}^f$). Dots in different colors represent instances from different domains. Notice that vanilla coarse2fine model shares the utterance representations in both stages and has only one matrix to visualize.}
\label{fig:cluster}
\end{figure}

One of our key innovations is the domain discrimination component, which can drive utterance representations to become consistent or distinguishable among multiple domains in the coarse stage or the fine stage, respectively. To validate this component, we map utterance representations from different domains into the 2-dimension space as Figure~\ref{fig:cluster} shows. The distribution of dots in different colors represents instances from multiple domains. The more confusing these dots are, the more domain invariant representations are. To compare these distributions quantitatively, we use Cali{\'n}ski-Harabasz~(CH)~\cite{calinski1974dendrite} to evaluate the clustering effect. Higher CH means more distinguishable distributions across different domains.

As shown in Figure~\ref{fig:cluster}, the coarse2fine model has a certain ability to distinguish instances from different domains, but still has difficulties to distinguish several domains in the middle. Although coarse2fine decodes domain invariant sketches and domain specific tokens in different stages, it shares utterance representations for decoding in both stages. And there are no explicit mechanisms to drive the model to learn focuses in different stages, especially given a small number of target instances. For our DAMP, in the coarse stage, dots in different colors tend to be mixed together, {hard to distinguish,} and CH score drops to 88.89, indicating much better domain general representations. In the fine stage, these dots tend to cluster by their colors, with a much higher CH, implying more distinguishable representations across different domains.
In DAMP, despite two encoders for the coarse and fine stages, the adversarial training component pushes the domain general representations hard to distinguish while making the fine-stage representations easier to classify.
This comparison shows again that the domain discrimination component does enable us to acquire domain invariant or domain specific utterance representations in different stages.

\subsection{Domain Relevance Attention}

Our domain relevance attention mechanism is designed to help the decoders concentrate on different aspects of utterances in different stages. Here we use a case study to show why this mechanism can actually help.

\begin{figure}
\includegraphics[width=0.41\textwidth]{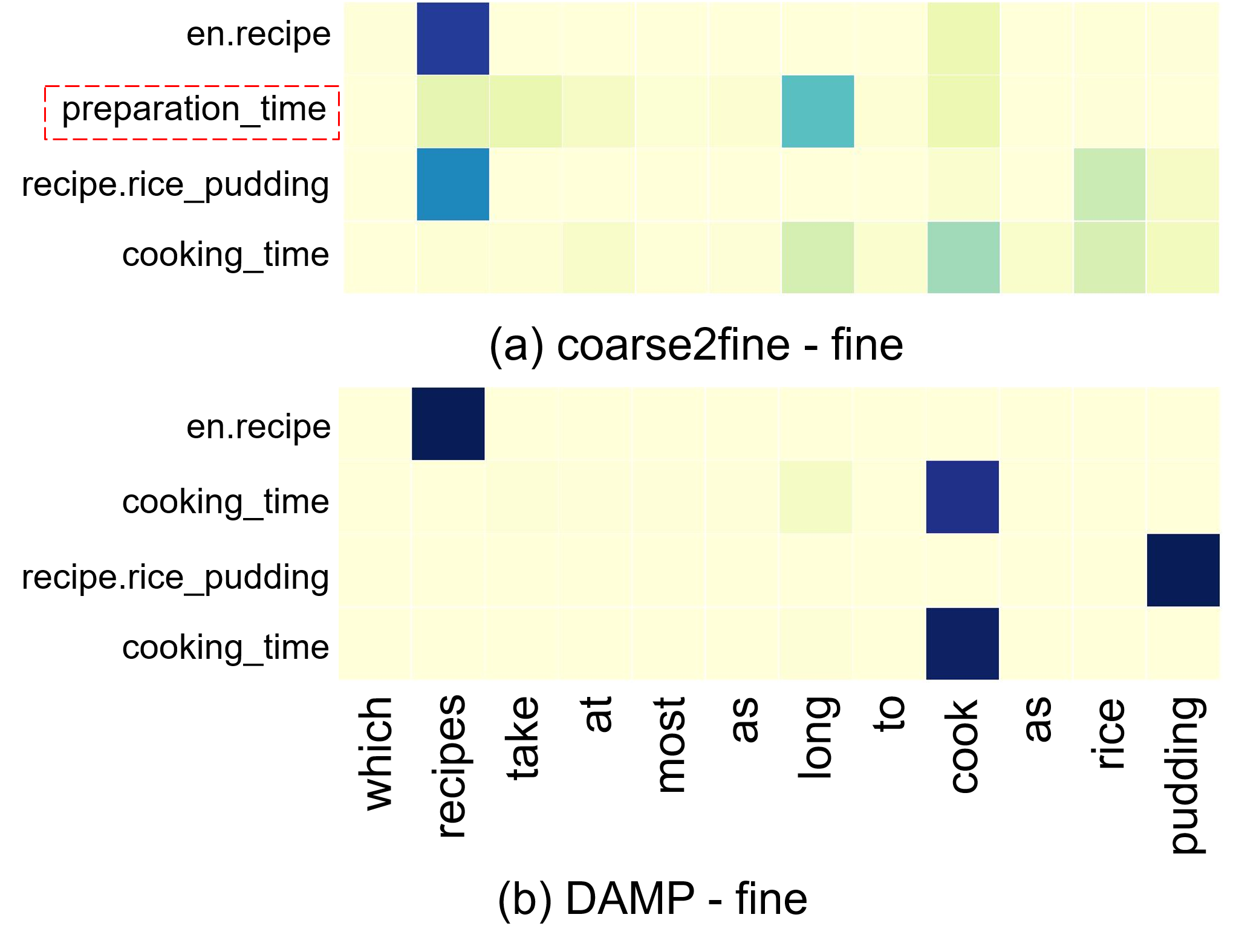}
\caption{Visualization of the attention distributions in the fine-stage decoder (decoder2 in Figure~\ref{fig:model}), which should generate 4 domain specific tokens (the 4 underlined positions) for the sketch  {\scriptsize\textsf{listValue (filter (getProperty (\underline{singleton@1}) (string !type)) (ensureNumericProperty (\underline{string@1})) (string\textless=) (ensureNumericEntity (\underline{getProperty@1} (\underline{string@1}))))}}, produced in the coarse stage. Darker colors represent larger attention scores. 
The coarse2fine model predicts a wrong domain-specific token  {\scriptsize\textsf{preparation\_time}} (in the red dotted box).
}
\label{fig:attention}
\end{figure}

Given the example sentence in Figure~\ref{fig:attention}, in the coarse stage, both models generate the right sketch with 4 domain specific positions to be filled in the fine stage. For the second position, coarse2fine splits its attention to \textit{recipes}, \textit{take}, \textit{at}, \textit{long} and \textit{cook}, while our DAMP only focuses on \textit{long} and \textit{cook}, with a majority to \textit{cook}. We can see that coarse2fine attends to both domain general keywords (e.g., \textit{at} and \textit{take}) and domain specific words, but our DAMP concentrates on the clue word \textit{cook} only. This difference makes coarse2fine incorrectly predict  \textsf{preparation\_time}, while our model outputs the right one, \textsf{cooking\_time}.    
Actually, it is the domain relevance attention mechanism that enables our model to focus on domain invariant patterns in the coarse stage, and concentrate on domain specific details in the fine stage. 

\subsection{More Tough Settings}
Regarding the model robustness, we evaluate our DAMP as well as three popular transfer strategies with different amounts of target domain training instances, from 1\% of the training data to 40\%. 

\begin{figure}
\centering
\includegraphics[width=0.47\textwidth]{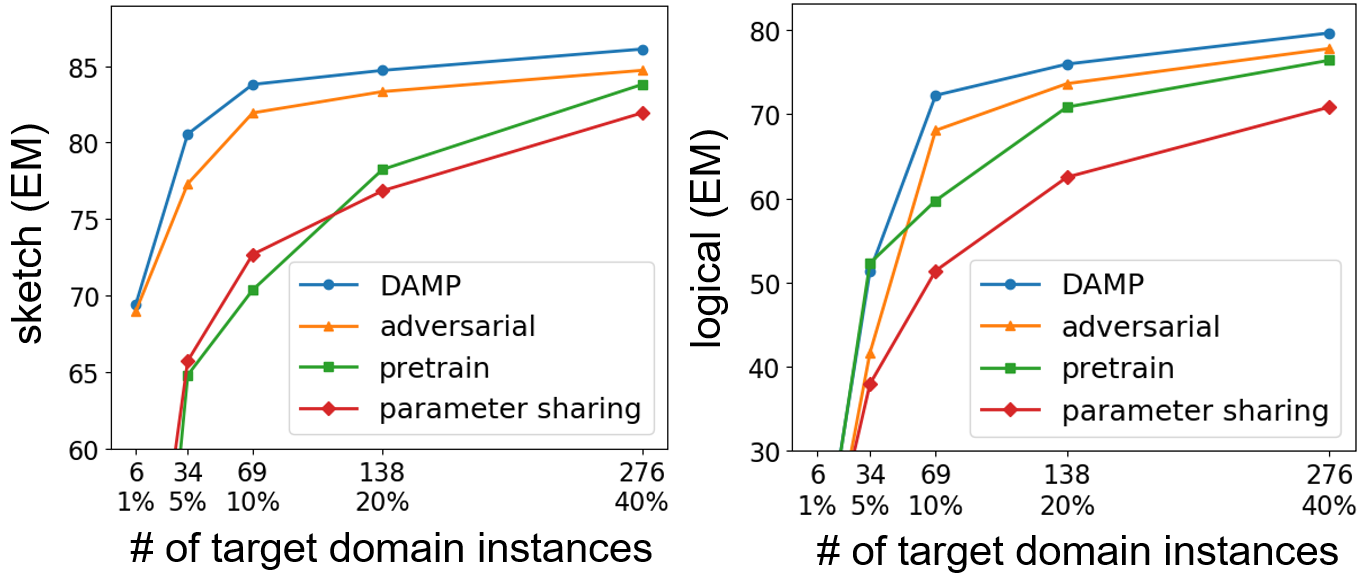}
\caption{ EM rate of the sketch (left) and the logical form (right) with different amounts of training data in the \texttt{recipes} domain. }
\label{fig:line}
\end{figure}

As illustrated in Figure~\ref{fig:line}, DAMP consistently outperforms all other transfer methods. With the increase of target domain training instances, DAMP curves quickly to go up for both sketch and logical form prediction. In other words,  DAMP can achieve the same performance with much fewer target domain training data,  which confirms the effectiveness of DAMP to leverage limited training data  in both stages.

We also notice DAMP can achieve as high as 69\% of sketch EM  with only 6 target domain instances. With 34 instances, it gets over 80\%. The reason may be DAMP can better exploit target instances to capture domain invariant representations through our adversarial domain discrimination, while other methods, e.g., pretraining~\cite{kennardi2019domain} and parameter sharing~\cite{liu2016recurrent}, are not good at distinguishing domain general and  specific information, especially with limited target data.
Looking at the performance for logical forms, we see all models perform badly with less than 34 target instances, since it is challenging to learn domain specific details with such a small amount of data. With a little more target instances, our model can achieve over 70\% EM of logical forms,    
because of its capability to focus on limited domain specific clues through our domain relevance attention.

\section{Related Work}

Recently, there have been increasing efforts using neural networks to solve semantic parsing tasks. Among many attempts, Sequence-to-sequence (seq2seq)~\cite{sutskever2014sequence} is a widely used framework for end-to-end solutions~\cite{jia2016data,zhong2017seq2sql}, where further efforts  take advantage of various structural knowledge, such as tree structures in logical forms~\cite{dong2016language} and semantic graph  \cite{chen2018sequence}. However, they do not separate the common patterns directly. \citet{dong2018coarse} proposed a two-stage semantic parsing method to separate the high-level skeletons from low-level details. In this way, decoders are able to model semantic at different levels of granularity. In this paper, we follow this two-stage framework, but explore its potential in the domain adaptation scenario. Particularly, we develop a novel domain discrimination component in the encoder and a domain relevance attention mechanism in the decoder, which separate domain specific information from general querying representations.

Domain adaptation is a long standing topic in the natural language processing community \cite{jiang2007instance,wu-etal-2017-active,wang2019adversarial}.
For semantic parsing, \citet{fan2017transfer} takes the multi-task setup to aids the transfer learning. And most recent works concentrate on how to bridge the gap of semantic representations between target and source domains. \citet{su2017cross} paraphrase utterance into the canonical forms, and utilize the word embedding to alleviate vocabulary mismatch between domains. 
\citet{lu2019look} alleviate the data scarcity by looking up a similar instance in memory of the target domain, and modifying it to the required logical form. 
However, it requires a large number of handcrafted grammars to distinguish  specific tokens.
\citet{xiong2019transferable} take a two-step process as well. But separation is not enough. We also develop mechanisms to emphasize the domain general and domain related information in two stages.

\section{Conclusions}

In this paper, we propose a new semantic parser, DAMP, for domain adaptation, which features two novel mechanisms to exploit limited target training data and focus on different clues at different stages.
Specifically, we propose an adversarial domain discrimination mechanism to learn domain invariant and domain specific representations in the coarse and fine stages, respectively, and design a domain relevance attention to drive decoders to concentrate on domain specific details. Detailed experiments on a benchmark dataset show that our model can well exploit limited target data and outperform widely adopted domain adaptation strategies.

\section*{Acknowledgments}

 We thank anonymous reviewers for their valuable suggestions. 
This work is supported in part by the National Hi-Tech R\&D Program of China (2018YFC0831900) and the NSFC Grants (No.61672057, 61672058). For any correspondence, please contact Yansong Feng.

\bibliographystyle{named}
\bibliography{ijcai20}

\end{document}